\documentclass[letterpaper, 10 pt, conference]{ieeeconf}  
\IEEEoverridecommandlockouts                              
\usepackage{times}

\usepackage{multicol}
\usepackage{graphics}
\usepackage[bookmarks=true]{hyperref}
\usepackage{svg}
\usepackage{amsmath,amssymb,bm}
\usepackage{setspace}
\usepackage[Symbol]{upgreek}
\usepackage{graphicx}
\usepackage{floatrow} 
\usepackage{multirow}
\usepackage{array}
\usepackage{enumerate}
\usepackage{sidecap}
\usepackage[all]{xy}
\usepackage{authblk}
\usepackage{comment}
\usepackage{changepage}   
\usepackage{wrapfig}
\usepackage{subfig}
\usepackage{caption}
\usepackage{url}
\usepackage{tabularx,pbox,booktabs}
\usepackage{flushend}
\usepackage{anyfontsize}
\usepackage{float}
\usepackage{todonotes}
\usepackage[percent]{overpic}
\graphicspath{{./pics/}}
\usepackage{amssymb}
\usepackage{dblfloatfix}    
\usepackage{booktabs}
\usepackage{amsfonts}
\usepackage[T1]{fontenc}

\usepackage{mathabx}
\usepackage{upgreek}
\usepackage{algpseudocode}
\usepackage{algorithm}

\usepackage{amsthm}

\algnewcommand\algorithmicforeach{\textbf{for each}}

\algdef{S}[FOR]{ForEach}[1]{\algorithmicforeach\ #1\ \algorithmicdo}
\algnewcommand\algorithmicinput{\textbf{Input:}}
\algnewcommand\INPUT{\item[\algorithmicinput]}
\algnewcommand\algorithmicoutput{\textbf{Output:}}
\algnewcommand\OUTPUT{\item[\algorithmicoutput]}

\title{Online Planning in Uncertain and Dynamic Environment\\ in the Presence of Multiple Mobile Vehicles}
\author{Junhong Xu, Kai Yin, Lantao Liu
\thanks{J. Xu and L. Liu are with the Luddy School of Informatics, Computing, and Engineering  at Indiana University, Bloomington, IN 47408, USA. E-mail:
        {\tt\small \{xu14, lantao\}@iu.edu}.
K. Yin is with Expedia Group. E-mail:
        {\tt\small kyin@expediagroup.com}.  
}
}

\begin{document}
\maketitle

\begin{abstract}
We investigate the autonomous navigation of a mobile robot in the presence of other moving vehicles under time-varying uncertain environmental disturbances.
We first predict the future state distributions of other vehicles to account for their uncertain behaviors affected by the time-varying disturbances.
We then construct a dynamic-obstacle-aware reachable space that contains states with high probabilities to be reached by the robot, within which the optimal policy is searched.
Since, in general, the dynamics of both the vehicle and the environmental disturbances are nonlinear, we utilize a nonlinear Gaussian filter -- the unscented transform -- to approximate the future state distributions. 
Finally,  the forward reachable space computation and backward policy search are iterated until convergence. 
Extensive simulation evaluations have revealed significant advantages of this proposed method in terms of computation time, decision accuracy, and planning reliability. 

\end{abstract}

\section{Introduction}

It is challenging for an autonomous robot to make decisions in a dynamic environment in the presence of other moving vehicles. 
Decisions of the robot must be computed fast to cope with uncertain or disrupting behaviors of other vehicles (either autonomous or non-autonomous, but are not under our control).
To overcome this challenge, 
the robot planning requires certain look-ahead knowledge of the dynamics for both our robot and other vehicles. 
However, the future states of other vehicles cannot be well predicted since they are uncontrollable to us (in fact, the future states of our robot cannot be estimated accurately either due to its uncertain motions/actions). 
Since the state distributions of both our controllable robot and other uncontrollable vehicles can be dependent on the environment which is oftentimes varying spatially and temporally (e.g., exotic disturbances such as time-varying winds or fluids that can perturb vehicle motion and decision), 
thus the time-varying stochasticity has to be incorporated in designing the planning mechanism.

Markov Decision Processes (MDPs)~\cite{puterman2014markov} 
have been widely utilized to formulate robotic decision-theoretic planning problems under uncertainty. 
Adding time-varying property usually induces higher computational demand for computing decisions~\cite{boyan2001exact}. 
We recently tackled this challenge by formulating the problem as a time-varying Markov Decision Process (TVMDP) and developing solutions that carefully exploit the state reachability characteristics~\cite{liu2018solution,XuYinLiu2019}. 
Unfortunately, the basic form of this method 
requires iterative computations to estimate the probabilistic distribution of look-ahead state transition time for each state which demand a considerable time complexity~\cite{XuYinLiu2019,liu2018solution}, leading to limited use for online planning problems.

%

In contrast to our prior solutions which essentially focused on estimating the look-ahead {\em time distributions} of state transitions, 
in this work we tackle the challenge from a totally different perspective: we will estimate the look-ahead {\em state distributions} instead of the {\em time distributions}.
In other words, the robot policies will be sought through explicitly modeling the time dependent process  of the state distributions for both our robot and other vehicles. 

The proposed new method achieves in planning in an online manner. 
It consists of a forward prediction step and a backward improvement/optimization step. 
Specifically, 
in the forward step we predict dynamic-obstacle-aware state distributions 
using a nonlinear Gaussian filter, the unscented transform~\cite{sarkka2013bayesian}. 
The state distributions allow us to construct the most reachable state space that contains states with high probabilities to be reached by the robot, 
within which the optimal policy can be searched.
In the backward step, we improve the policy based on the results from the prediction step. 
The policy is optimized by performing the forward prediction and backward policy search iteratively, 
which maximizes the long-term return of the robot
and, at the same time, considers stochastic behaviors of other vehicles due to time-varying disturbances.

This paper includes the following contributions:
\begin{itemize}
     \item To solve the underlying TVMDP, we propose a time-discretization based solution to estimate robot and vehicles' future state distributions. This allows us to avoid cumbersome and iterative state transiting time estimation, leading to significantly improved time complexity.
    \item To further mitigate the computation, we propose an algorithm to construct the most reachable state space based on the bounds of look-ahead state distributions. 
     \item We design a fast online policy search algorithm within the space of high reachablility to solve the planning problem in the presence of other moving vehicles with uncertain behaviors.  
\end{itemize}

\section{Related Work}

Planning in dynamic and uncertain environments in the presence of moving vehicles can be modeled as decision-theoretic planning~\cite{boutilier1999decision}.
Typical existing methods~\cite{galceran2015multipolicy, mehta2018backprop, cunningham2015mpdm} formulate this problem as Partially Observable Markov Decision Process (POMDP) where the behavior of other vehicles are not observable but assumed to be selected from a fixed number of closed-loop policies.
The deterministic rollouts are then used to determine the best policy to execute.
A similar work~\cite{bandyopadhyay2013intention} models this problem as a mixed observability MDP, which is a variant of POMDP~\cite{kurniawati2008sarsop}. 
A more general framework is proposed in~\cite{du2011robot} where the authors combine motion prediction and receding horizon planning to reduce the uncertainty during planning. 

In addition to planning methods, learning-based approaches can also be used to deal with dynamic environments.
For instance, reinforcement learning has been used to learn navigation policy in social environments~\cite{chen2017socially, kretzschmar2016socially}.
Probabilistic inference methods, specifically Gaussian Processes, have also been used to predict the behaviors of moving agents and perform planning based on the predictions~\cite{joseph2011bayesian, trautman2010unfreezing}.

Although the aforementioned methods take environmental uncertainty into account, they do not deal with time-varying stochasticity. 
To account for time-varying uncertainty, our previous work~\cite{liu2018solution,XuYinLiu2019} develop approximate solutions to time-varying Markov Decision Processes.
The time variability has been used to restrict the policy search space~\cite{XuYinLiu2019}.
Unlike the previous work, we exploit distribution over spatial states to construct the reachable space. 
This idea is also related to policy search methods~\cite{kober2009policy, deisenroth2013survey, levine2014learning} which iteratively search for a local control policy.

Proximal work also includes trajectory optimization methods which utilize iterative mechanisms  to find local policies with rollout computation. 
For example, the forward and backward passes have been used for policy computation~\cite{tedrake2010lqr,tassa2012synthesis}. 
Similar methods can also be found in~\cite{tassa2014control}, where the differential  dynamic programming is employed to calculate policies under control constrains. 
In general, these methods are either based on sampling or deterministic rollouts, which is different from our proposed method that directly bounds the search space with a time discretization scheme.

\section{Preliminaries and Problem Formulation}
We first introduce the general form of the decision-making problem in time-varying  environments.
Then, we formulate this problem as a TVMDP, which allows us to develop the online decision-making algorithm in a principled manner. 

\subsection{Decision-Making with Time-Varying Uncertainties}\label{sec:decision-problem}

\subsubsection{Robot motion} 
we formulate the robot motion as a discrete-time nonlinear dynamical system with time-varying additive external disturbance
\begin{equation}\label{eq:dynamics}
    \mathbf{x}_{k+1} =  f(\mathbf{x}_k, \mathbf{u}_k) + e(\mathbf{x}_k, t_{k}),
\end{equation}
where states $\mathbf{x} \in \mathbb{R}^{D}$ and controls $\mathbf{u} \in \mathbb{R}^{M}$ are continuous multi-dimensional variables with $D$ and $M$ elements; 
the index $k \in \mathbb{Z}_{+}$ denotes the discrete decision step; 
$\mathbb{T}$ is a countable set that contains real-valued discrete decision times with equal interval $\Delta t$, i.e, $\mathbb{T} = \{t_0, t_1, ..., t_k, ...\}$ with $t_k = t_0 + k\Delta t$;
$f(\cdot, \cdot)$ describes the dynamical model of the robot. 
The noisy disturbance term $e(\mathbf{x}_k, t_{k}) = g(\mathbf{x}_k, t_{k}) + \epsilon_k$ captures deterministic time-dependent external disturbances $g(\cdot, \cdot)$ and random noises 
$\epsilon_k$.
We assume that the control $\mathbf{u}_k$ is applied for a period of $\Delta t$ time at each decision step. 
Equation~(\ref{eq:dynamics}) allows us to 
derive the conditional distribution of states when the probability density function of $\epsilon_k$ has a closed form. In particular, if
$\epsilon_k \sim \mathcal{N}(\mathbf{0}, \mathbf{Q}_k)$, where $\bf{Q}_k$ is the covariance of the noise term, then 
\begin{equation}\label{eq:continuious-state-distribution}
  \mathbf{x}_{k+1} \sim q(\mathbf{x}_{k+1} | \mathbf{x}_k, \mathbf{u}_k, t_k) := \mathcal{N}(\mathbf{x}_{k+1} | f + g, \mathbf{Q}_k).
\end{equation}

\subsubsection{Robot decision process} 
\label{sec:robot-decision-process}

the robot decision is described by a deterministic policy $\pi^c: \mathbb{R}^{D} \times \mathbb{T} \rightarrow \mathbb{R}^{M}$ which maps a continuous state and a decision time to a continuous action $\mathbf{u}_k$.
The expected total reward with a starting state $\mathbf{x}_0$ from the first decision step $t_0$ under $\pi^c$ is represented by
 $   v^{\pi^c}(\mathbf{x}_0, t_0) = 
    \sum_{k=0}^{\infty}\int_{\mathbb{R}^{D}} q(\mathbf{x}_k)r(\mathbf{x}_k, \mathbf{u}_k, t_k)d\mathbf{x}_k,$
where $q(\mathbf{x}_k)$ is the probability density of $\mathbf{x}_k$, $\mathbf{u}_k = \pi^c(\mathbf{x}_k, t_k)$, and $r(\mathbf{x}_k, \mathbf{u}_k, t_k)$ is a one-step look-ahead reward function which implicitly depends on the other vehicles at $t_k$. 
We aim to search for a policy $\pi^{c*}$ that maximizes the expected total reward. 

\subsubsection{Motions of other dynamic vehicles} 
we assume that the motion of each $i^{th}$ uncontrollable vehicle is also disturbed by the external disturbance and is described by
\begin{equation} \label{eq:othervehicle}
    \mathbf{y}_{k+1}^{i} = z^i(\mathbf{y}_k^{i}, \mathbf{u}_{k, y}^i) =  
    f^i(\mathbf{y}^i_{k}, \mathbf{u}_{k, y}^i) + e(\mathbf{y}_k^i, t_{k}),
\end{equation}
where $\mathbf{y}^i \in \mathbb{R}^{D_i}$ is the state, $\mathbf{u}_{y}^{i} \in \mathbb{R}^{M_i}$ is the decision generated by the $i^{th}$ vehicle's decision process, and $f^i(\cdot, \cdot)$ represents the motion model of the $i^{th}$ uncontrollable vehicle. 
We assume that their states are observable, but the policies corresponding to the decision processes (i.e., $\mathbf{u}_{y}^{i}$) of these vehicles are unknown and can only be estimated. 
In this work, we use the social force model (SFM)~\cite{helbing1995social} to model, and approximate, their behaviors.
Therefore, the exact future states of the uncontrollable vehicles cannot be predicted with certainty because their motions are under the time-varying uncertain disturbance and the predictions of their decisions are also uncertain.

\subsection{Time-Varying Markov Decision Processes}

To solve the decision problem introduced in Section~\ref{sec:decision-problem}, we model it as a discrete-time time-varying Markov Decision Process (TVMDP) \cite{XuYinLiu2019}. Similar to the methods used in \cite{gorodetsky2018high} and \cite{chow1991optimal}, 
we partition the continuous state and action spaces into subareas for discretization, and each subarea corresponds to a discrete state or action. 
The obtained TVMDP is thus represented as a 5-tuple 
$(\mathbb{S}, \mathbb{T}, \mathbb{A}, \mathcal{T}, r)$, 
where $s \in \mathbb{S}\subset \mathbb{R}^D$ and $a \in \mathbb{A}\subset \mathbb{R}^M$ are the discrete states and actions, respectively. 
To map a continuous state to a discrete state, we define the function $s = \psi_s(\mathbf{x})$ for such a purpose.
Similarly, a continuous action can be mapped to its discrete counterpart by $a = \psi_a(\mathbf{u})$. 

Policies are now mappings from discrete states and time to discrete actions $\pi: \mathbb{S} \times \mathbb{T} \rightarrow \mathbb{A}$. 
The state transition model from state $s'$ to $s$ with action $a$ is written as $\mathcal{T}_a(s, s', t) = p(s' | s, a, t)$. 
According to the robot dynamics Eq.~\eqref{eq:continuious-state-distribution}, 
the transition function 
is defined as 
\begin{equation}\label{eq:discrete-transition}
   \mathcal{T}_{a_{k}}(s_{k+1}, s_k, t_k)  = \frac{\int_{\psi^{-1}_s(s_{k+1})}q(\mathbf{x} | s_k, a_k, t_k)d\mathbf{x}}{\sum_{s' \in \mathbb{S}'}\int_{\psi^{-1}_s(s')}q(\mathbf{x} | s_k, a_k, t_k)d\mathbf{x}},
\end{equation}
where $\mathbb{S}'$ is the set of possible discrete states at $t_{k+1}$.

Similarly, the reward function becomes $r_a(s, s', t)$. 
The value function via Bellman equation yields
$v^{*}_{s, t_{k}} = \max_{a\in A} \sum_{{s'\in S}} \mathcal{T}_a( s, s', t_k) \Big( r_a(s, s', t_{k})  + \gamma ~v^{*}_{s', t_{k+1}} \Big)$,
where $v_{s, t_k}^*$ means the optimal value function of a state $s$ at time $t_k$, and $\gamma \in [0, 1)$ is a discount factor.


We recently designed a framework without discretizing time 
so the time complexity remains at the same level~\cite{liu2018solution}.
The key idea is to 
{\em evolve} the spatial state transitions along the temporal dimension where each state's stochastic transition time needs to be explicitly estimated. The spatial and temporal processes are coupled by taking advantage of the underlying vehicle dynamics. 


\vspace{-3pt}
\section{Methodology}
\begin{figure}[t]
    \centering
    \subfloat{\label{fig:k-0-action-selection} \includegraphics[width=0.95\linewidth]{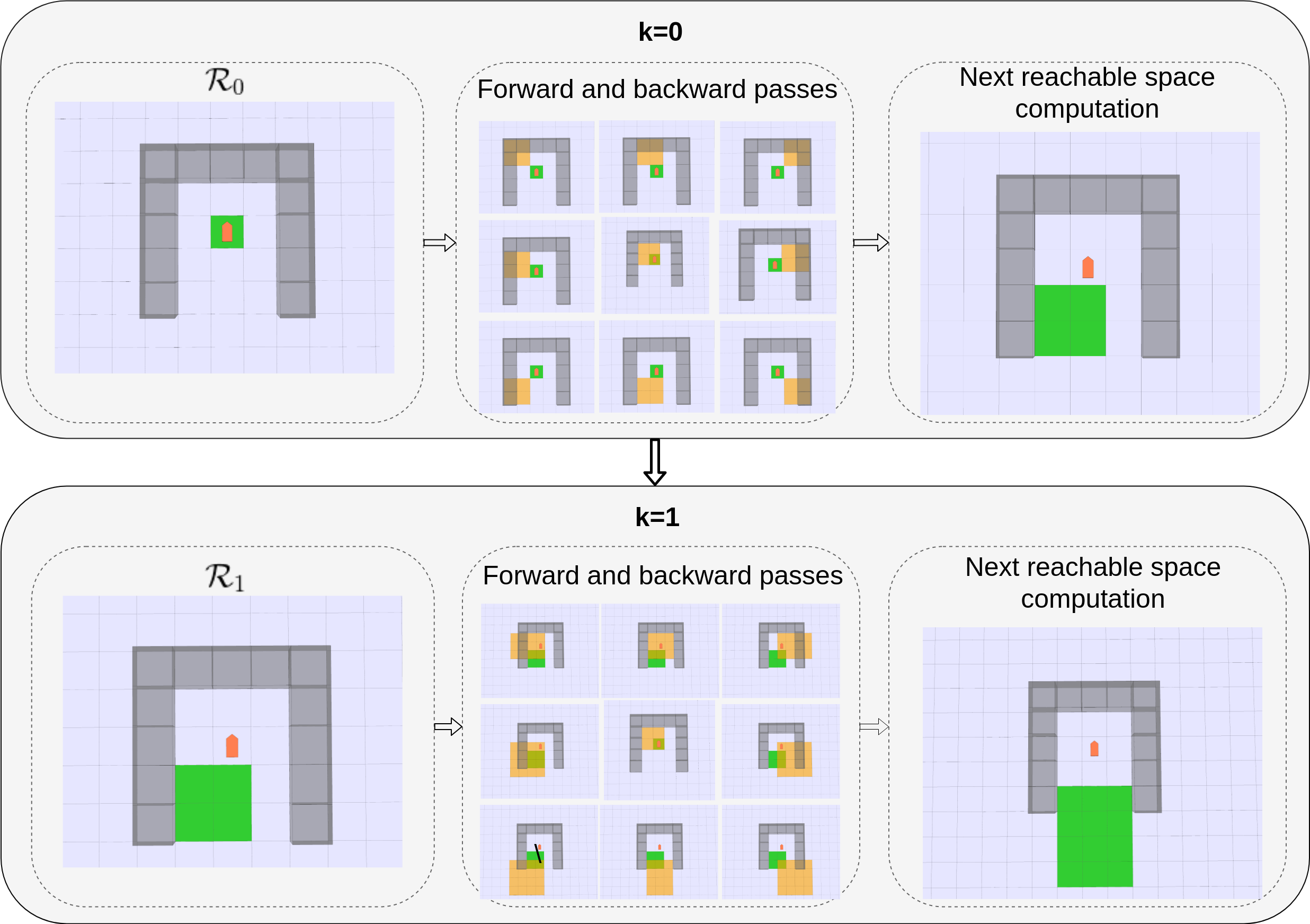}}
    \caption{\small  An illustrative example of Alg.~\ref{alg:fast-policy-search} at $k=0$ and $k=1$ decision step. The green grids represent the reachable space of current policy at the current decision step; the orange grids are the reachable spaces of each action.    \vspace{-15pt} 
    }    
    \label{fig:action-selection}
\end{figure}



The essence of this work is to design a computationally tractable online planning approach that incorporates uncertainties of the environment and the prediction of other vehicles' states.
To remove the computational barrier without deteriorating the accuracy, we construct a dynamic-obstacle-aware reachable space that contains the states with high probabilities to be reached by the robot. 
The reachable space construction is based on the state distributions of the robot and other vehicles in the  time-varying  environments. 
Due to the nonlinearity of the vehicle dynamics and the time-varying disturbances, we apply an efficient numerical integration scheme -- the nonlinear Gaussian filters via unscented transform~\cite{sarkka2013bayesian} -- to approximate the future state distributions for all vehicles.
Finally, a valid policy is searched within the reachable space.



\subsection{State Distribution Modeling based on Gaussian Filtering in Discretized Time Dimension}\label{discretization}


Different from the previous scheme which estimates {\em time distributions} based on discretized state space~\cite{XuYinLiu2019}, in this work we will estimate {\em state distributions} based on discretized time dimension. 
The spatial and temporal dimensions are treated separately so that we can exploit the structure of state reachability.
The spatial and temporal processes are coupled by the underlying real-world vehicle motion dynamics (Eq.~\eqref{eq:dynamics} and Eq.~\eqref{eq:othervehicle}) which are functions that describe spatial vehicle states with respect to time. 
The separation and unification of spatiotemporal spaces lead to great flexibility that facilitates the estimation of state distributions with time-varying stochasticity.  
(Note, the state distribution here is different from the generic POMDP's belief as all vehicles' states can be observed although future states need to be predicted.)



{
\begin{algorithm}[t]
{\small 
\caption{Reachable Space Based Online Policy Search}
\label{alg:fast-policy-search}
\begin{algorithmic}[]
    \INPUT{TVMDP elements $(\mathbb{S}, \mathbb{T}, \mathbb{A}, \mathcal{T}, R)$; planning horizon $T$; the starting state $\mathbf{x}_0$; starting time $t_0$; time $\Delta t$; the confidence level $\alpha$.}
    \OUTPUT{policy $\pi$}
    \State Initialize $\mathcal{R}_0^{\pi} = \{\mathbf{s}_0\}$, $\mu_0 = \mathbf{s}_0$, and $\Sigma_0 = \mathbf{0}$.
    \Repeat
        \For{$k=0, ..., T-1$}
            \State // Construction of reachable space for each action. 
            \State Compute $\mathcal{R}_{k+1}^A$ based on Alg.~\ref{alg:compute-reachable-space-greedy}.
            \State // Backward policy and value update.
            \For{$s \in \mathcal{R}_k^{\pi}$}
                \State Update $\pi(s, t_k)$ and $v_{s, t_k}$ 
                 using Eq.~(\ref{eq:mdp-t-ReachableSpace})  
            \EndFor
            \State // Forward reachable space construction.
            \State Compute $\mu_{k+1}^{\pi}$ and $\mathbf{\Sigma}_{k+1}^{\pi}$ based on Eq.~(\ref{eq:moment-matching-mean}) and Eq.~(\ref{eq:moment-matching-var}) under the updated policy $\pi$. 
            \State Find $\mathcal{R}_{k+1}^{\pi}$ using $\mu_{k+1}^\pi$, $\mathbf{\Sigma}_{k+1}^{\pi}$.
        \EndFor
    \Until The policy does not change or the algorithm reaches its time budget. 
  \end{algorithmic} \vspace{-1pt}
  }   
\end{algorithm} 
}





To estimate the state distributions, the nonlinearity of vehicle dynamics have to be considered.
We apply the nonlinear Gaussian filter, the unscented transform, for such prediction task.  
Formally, given a policy $\pi$, the state distribution at the decision step $k+1$ is computed based on the prediction of the current state distribution $q_{t_k}^{\pi}(\mathbf{x}_{k})$ 
\begin{align}\label{eq:belief}
   q_{t_{k+1}}^\pi(\mathbf{x}_{k+1}) &= 
   \int q_{t_k}^{\pi}(\mathbf{x}_{k})q(\mathbf{x}_{k+1}| \mathbf{x}_{k},
  \pi(x_k, t_k), t_k)d\mathbf{x}_k
\end{align}
where $q(\mathbf{x}_{k+1}| \mathbf{x}_{t},\pi(x_k, t_k), t_k)$ is given in Eq. (\ref{eq:continuious-state-distribution}), and $q_{t_k}^{\pi}(\mathbf{x}_k)$ is the probability density of $\mathbf{x}_k$ at the current decision step $t_k$ given policy $\pi$.
The integral on the right hand side of Eq.~\eqref{eq:belief} may be numerically computed with, for example, the Gauss-Hermite quadrature method \cite{hildebrand1987introduction}.
Then we can predict the moment estimation of the state $\mathbf{x}_{k+1}$.
Let us denote the mean and covariance of $\mathbf{x}_{k+1}$ at $t_{k+1}$ under policy $\pi$ by $\mu_{k+1}^{\pi}$ and $\mathbf{\Sigma}_{k+1}^{\pi}$, respectively.
Both can be obtained by 
the unscented transform approach~\cite{wan2000unscented} which belongs to the Gaussian filtering methods~\cite{sarkka2013bayesian} that numerically approximate the integral. We have
\begin{align}
    \label{eq:moment-matching-mean}
    \mu_{k+1}^{\pi} = \mathbb{E}_{t_{k+1}}^{\pi}[\mathbf{x}_{k+1}]&\approx
    \sum_{i=0}^{2n}W_i^{m}z^{\pi}(\mathbf{x}^{i}_k, t_k),\\ 
    \label{eq:moment-matching-var}
    \mathbf{\Sigma}_{k+1}^{\pi} = \mathbb{V}_{t_{k+1}}^{\pi}[\mathbf{x}_{k+1}] &\approx 
    \sum_{i=0}^{2n}W^{c}_i[z^{\pi}(\mathbf{x}^{i}_k, t_k) - \mu_{k+1}]\cdot\nonumber\\
    &[z^{\pi}(\mathbf{x}^{i}_k, t_k) - \mu_{k+1}]^T + \mathbf{Q}_k, 
\end{align}
where $z^{\pi}(\mathbf{x}^{i}_k, t_k):=f(\mathbf{x}_k^i, \pi(\mathbf{x}_k^i, t_k))+g(\mathbf{x}_k^i, t_k)$ is defined in Section~\ref{sec:decision-problem}; $W_{i}^{m}$, $W_i^{c}$ are the constant weights for approximating the mean and covariance; the points $\mathbf{x}_k^{i}$ are the so-called \textit{sigma points} which are selected deterministically based on the current mean and covariance. 

\subsection{Reachable Space based Online Policy Search}
\label{reachable-sapce-search}

\subsubsection{Reachable space computation}
The previously introduced unscented transform enables us to predict the spatial state distribution $q_{t_{k+1}}^{\pi}(\mathbf{x}_{k+1})$ along the temporal dimension.
As a result, we can compute the most probable space reached by the robot given a policy. We call such space \textit{the reachable space} $\mathcal{R}_{k+1}^{\pi}$, and we define it as the confidence region of $\mathbf{x}_{k+1}$. 
The confidence region of $\mathbf{x}_{k+1}$ is a D-dimensional ellipsoid centered at the mean of the distribution and its spread and direction are determined by the covariance matrix
\begin{equation}\label{eq:confidence-region}
   \{\mathbf{x} : (\mu_{k+1}^{\pi} - \mathbf{x})^T \Sigma_{k+1}^{\pi} (\mu_{k+1}^{\pi} - \mathbf{x}) \leq \mathcal{X}_D^2(\alpha)\}, 
\end{equation}
where $\alpha$ is the significance factor that determines the confidence level and $\mathcal{X}_D^2$ is the cumulative distribution function of the chi-squared distribution with $D$ degrees of freedom~\cite{bishop2006pattern}.
Thus Eq.~\eqref{eq:confidence-region} gives the confidence region over the continuous state space.
$\mathcal{R}_{k+1}^{\pi}$ is then found by including all the states within the boundary of the ellipsoid. 

\subsubsection{Policy search} 
The online policy search algorithm is illustrated in Alg.~\ref{alg:fast-policy-search}, which aims at calculating a valid policy within a limited time budget (generally less than $1$ second). This is achieved by constraining the search within the reachable space.
The policy search includes three major stages. 

The first stage constructs the reachable space for each action. 
Suppose the reachable space of the current policy $\pi$ at $t_k$ is $\mathcal{R}_k^{\pi}$ and the sigma points are $\mathbf{x}_k^i$ (see Section~\ref{discretization}). 
Assuming an action $a$ is taken at all $\mathbf{x}_k^i$, we then get the mean $\mu_{k+1}^a$ and covariance $\Sigma_{k+1}^a$ of next visited states by equations similar to Eq.~(\ref{eq:moment-matching-mean}) and ~(\ref{eq:moment-matching-var}). 
That is, we replace $\pi(\mathbf{x}_k^i, t_k)$ by $a$ for all $\mathbf{x}_k^i$ in Eq.~(\ref{eq:moment-matching-mean}) and ~(\ref{eq:moment-matching-var}) to compute the results. Then we can get the reachable space of action $a$ via Eq.~(\ref{eq:confidence-region}) using $\mu_{k+1}^a$ and $\Sigma_{k+1}^a$. 
Let the reachable space of action $a$ be $\mathcal{R}_{k+1}^a$, and $\mathcal{R}^A_{k+1} =\bigcup_{a \in \mathbb{A}} \mathcal{R}_{k+1}^a$. This first stage procedure is summarized in 
Alg.~\ref{alg:compute-reachable-space-greedy}.

The second stage constrains the Bellman equation to proceed only within the state space $\mathcal{R}^A_{k+1}$ to obtain the policy $\pi(s, t_k)$  and values $v_{s, t_k}$ for states  $s\in\mathcal{R}_k$, i.e.,
\begin{equation} \label{eq:mdp-t-ReachableSpace}
v_{s, t_k} = \max_{a\in A} \sum_{{s'\in \mathcal{R}^A_{k+1}}} \mathcal{T}_a(s, s', t_{k}) \Big(r_a(s, s', t_{k})  + \gamma v_{s', t_{k+1}} \Big),
\end{equation}
where $\pi(s, t_k)$ corresponds to the resulting action $a$.
As the reachable space contains states that are most likely reached by the vehicles, the policy improvement constrained within this space results in a solution equal to or close to the optimal.

Finally, in the third stage, the algorithm uses the updated policy to obtain the reachable space of policy $\pi$, i.e., $\mathcal{R}_{k+1}^{\pi}$. 
An illustration of the algorithm is shown in Fig.~\ref{fig:action-selection}. 

\begin{algorithm} [t]
{\small 
\caption{Reachable Space of Action}
\label{alg:compute-reachable-space-greedy}
  \begin{algorithmic}[]
    \INPUT{Confidence level $\alpha$; mean $\mu_k^{\pi}$; covariance matrix $\Sigma_k^{\pi}$; reachable space $\mathcal{R}_k^{\pi}$ and sigma points $\mathbf{x}_k^i$; discrete action space $\mathbb{A}$.}
    \OUTPUT{The set of states $\mathcal{R}^A_{k+1}$}.
    \State Initialize $\mathcal{R}^A_{k+1} = \emptyset$.
    \For{$a \in A$}
        \State Compute $\mu$ and $\Sigma$ based on Eq.~(\ref{eq:moment-matching-mean}) and Eq.~(\ref{eq:moment-matching-var}) with replacing $\pi(\mathbf{x}_k^i, t_k)$ by $a$ as described in Section~\ref{reachable-sapce-search}.
        \State $\mathcal{R}_{k+1}^a:=\{\mbox{the set by Eq.~(\ref{eq:confidence-region}) using }\mu, \Sigma, \alpha\}$.
        \State $\mathcal{R}^A_{k+1} = \mathcal{R}^A_{k+1} \bigcup\mathcal{R}^a_{k+1}$.
    \EndFor
  \end{algorithmic}
  }
\end{algorithm}

\subsection{Reward Function and State Prediction of Other Vehicles}\label{sec:agent-prediction}
\begin{figure}\vspace{-8pt}
    \centering
    \subfloat[$k=1$]{\label{fig:}\includegraphics[width=0.47\linewidth]{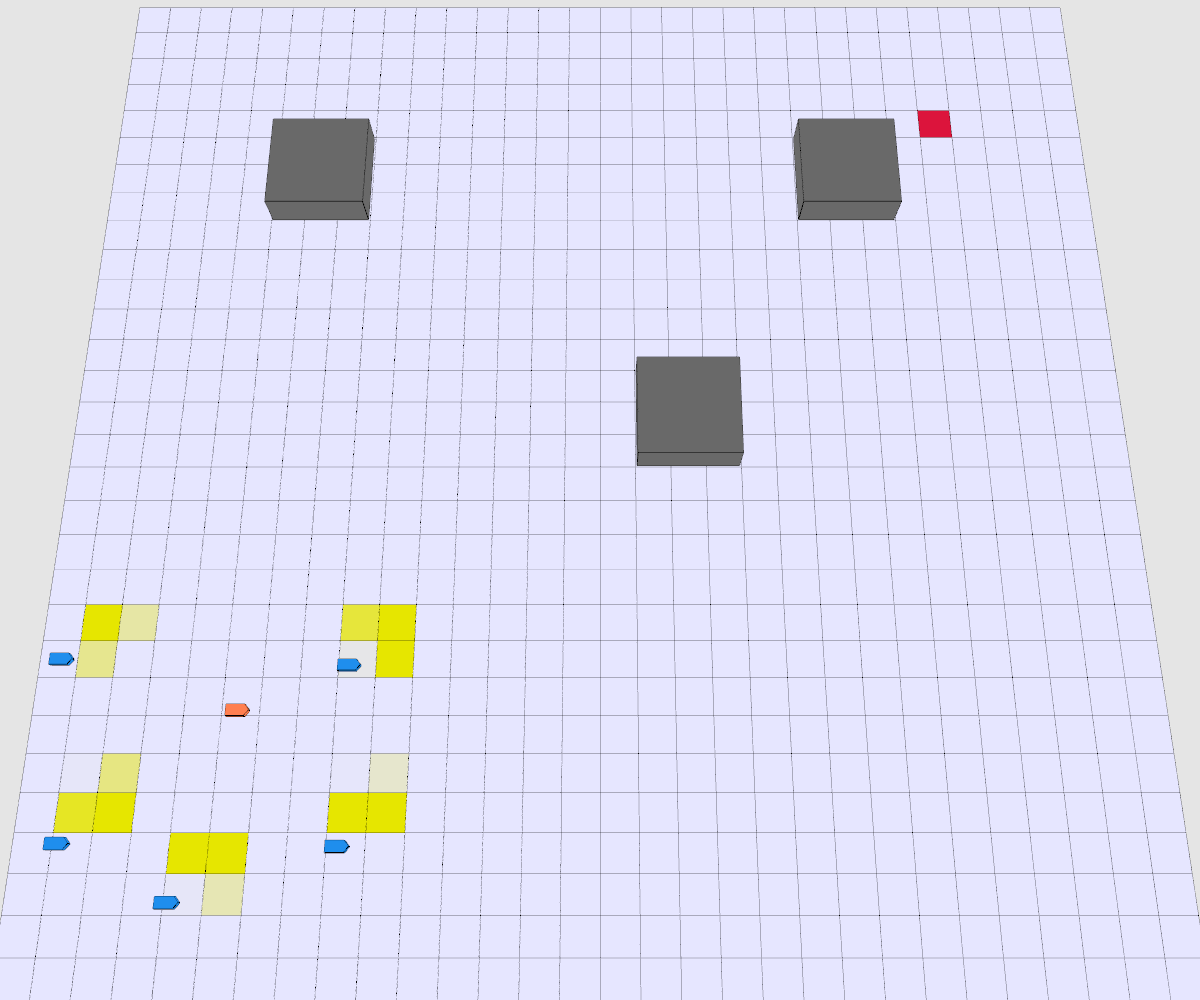}}\quad
    \subfloat[$k=4$]{\label{fig:}\includegraphics[width=0.47\linewidth]{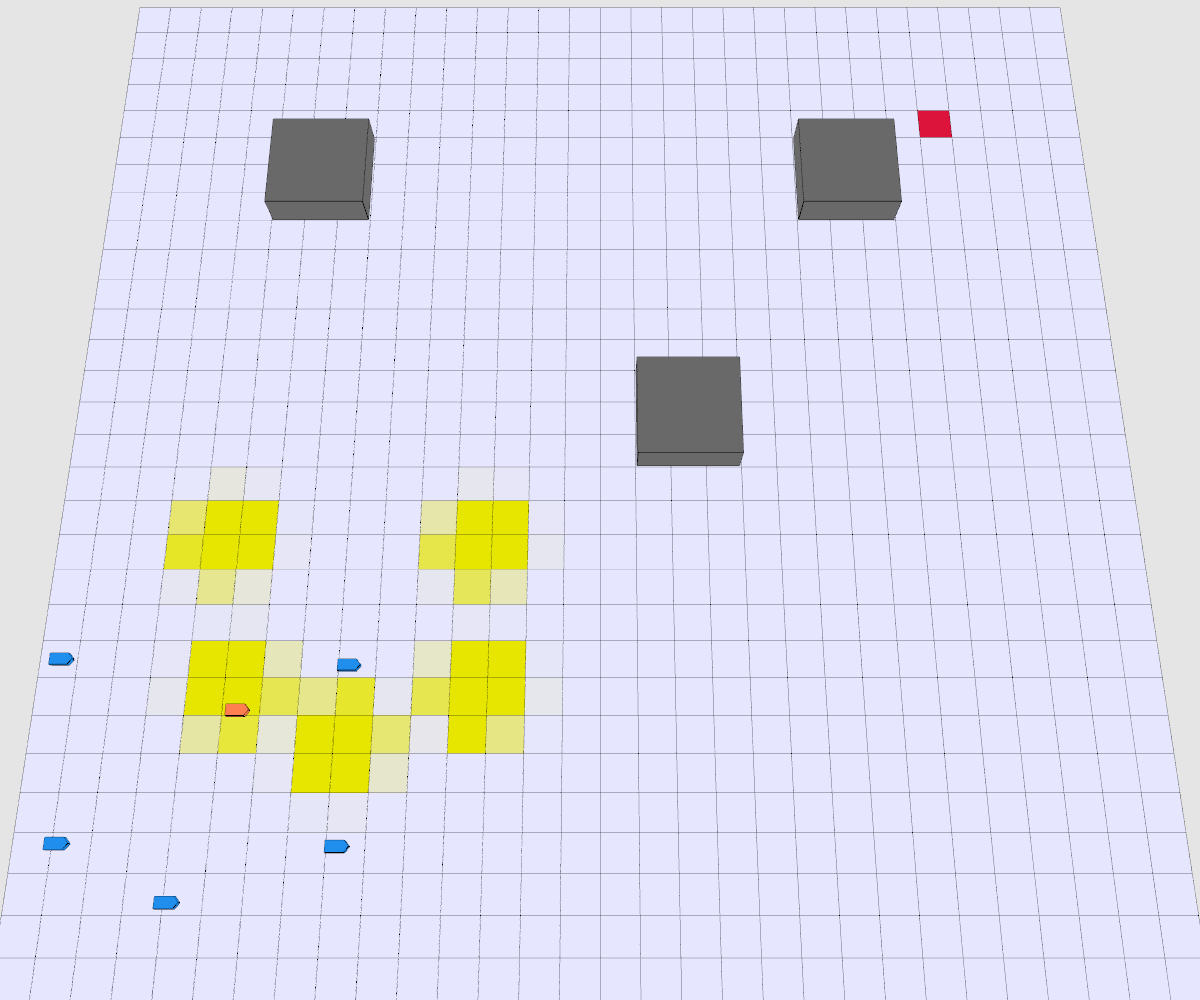}}
    \caption{\small An example of motion predictions of other agents. (a)(b) show the look-ahead predictions at $k=1$ and $k=4$, respectively.
    The orange and blue vehicles are the controllable and uncontrollable agents, respectively. 
    The yellow cells are the predicted positions of other vehicles. 
    Color intensity indicates the certainty level of the prediction.     
    \vspace{-15pt}
    }
    \label{fig:motion-prediction}
\end{figure}

The reward is a function that relates to vehicle future states.
To model the collective behavior of other vehicles,  
we opt to use the Social Force Model (SFM)~\cite{helbing1995social} to capture the responses among vehicles based on which their future states can be predicted. 
The SFM is a behavioral model that describes the interaction phenomena among mobile agents, and the underlying rules fit well for vehicles in space-limited environments, e.g., if two vehicles are too close to each other, they tend to separate to be safe. 
(Note that, the collective behaviors for other uncontrollable vehicles can be different in various scenarios and applications. Here we use the SFM to demonstrate the basic idea for calculating the reward. We believe other given or known behavioral rules are directly applicable too.)

Specifically, for the $i^{th}$ uncontrollable vehicle, the action computed by the SFM at time step $k$ is denoted by $\mathbf{u}_{k,y}^{i}$. 
The state distribution of the mobile vehicles at $t_k$ can then be computed based on the states at $t_{k-1}$, namely,
\begin{align}\label{eq:motion-prediction-uncontrollable}
    q(\mathbf{y}_{k}^i) = \int &q(\mathbf{y}^i_k | \mathbf{y}^i_{k-1}, \mathbf{u}^i_{k-1, y}, t_{k-1})\\ 
    & q(\mathbf{x}_{k-1}, \mathbf{y}_{k-1}^{1, ..., N})d\mathbf{x_{k-1}}d\mathbf{y}_{k-1}^{1, ..., N},\nonumber
\end{align}
where $\mathbf{y}^{i}$ is the state of the $i^{th}$ vehicle, $\mathbf{y}^{1, ..., N}$ is the vector of states of all the mobile vehicles, and $\mathbf{u}^i_{k-1, y}$ is given by SFM.
Then, the state predictions of other vehicles can be solved using the unscented transform approach presented in Section~\ref{discretization}. 

Then we construct the reward function based on the state predictions of other vehicles.
The reward should discourage the collision between the robot and other vehicles; it is given by the following equation
\begin{equation}
   r_{a_k}(s_k,t_k) = \sum_{s' \in \mathcal{R}_{k+1}}{\mathcal{T}_{a_k}(s', s_k, t_k)r_{a_k}(s_k,  s', t_k)}, 
\end{equation}
where 
$r_{a_k}(s_k, s', t_k) = \eta \sum_{i=1}^N p(\mathbf{y}_{k+1}^i = {s}')\cdot c$. Here $c$ is the collision penalty;
coefficient $\eta$ is a normalizing factor; $p(\mathbf{y}^i_{k+1} =s_{k+1})$ is the probability of the $i^{th}$ vehicle arriving at the next state $s'$. 
An illustration of the predicted state distributions of 5 agents is shown in Fig.~\ref{fig:motion-prediction}.

\subsection{Online Planning Algorithm}
The algorithm is presented in Alg.~\ref{alg:online-planning}. 
Briefly, the robot first observes the current states of other vehicles, 
then the reward function is calculated by predicting the state distributions of other uncontrollable vehicles. 
With that, Alg.~\ref{alg:fast-policy-search} is employed 
to compute a policy for the robot to execute in a receding horizon manner.


\begin{algorithm}[t] 
{\small 
\caption{Online Planning in Dynamic Environments}
\label{alg:online-planning}
  \begin{algorithmic}[]
    \INPUT{Planning time interval $\tau$; 
          confidence level $\alpha$;
          initial continuous state $\mathbf{x}_0$;
          initial time $t_0$; time $\Delta t$.}
    \State Initialize $t = t_0$, $\mathbf{x} = \mathbf{x}_0$, and $T = \frac{\tau}{\Delta t}$.
    \Repeat
        \State Observe other agent states $\mathbf{y}^1, ..., \mathbf{y}^N$.
        \State Compute reward function $r$ in Section~\ref{sec:agent-prediction}.
        \State Compute $\pi$ based on Alg.~\ref{alg:fast-policy-search}.
        \For{$k$ in $0 ... T-1$}
            \State $\mathbf{u} = \pi(\psi_s(\mathbf{x}), t)$.
            \State Apply the action $\mathbf{u}$ for $\Delta t$ seconds and observe a new state $\mathbf{x}$.
            \State $t = t + \Delta t$.
        \EndFor
    \Until The goal is reached. 
  \end{algorithmic}
  }
\end{algorithm}

\vspace{-15pt}
\section{Experiments}

\begin{figure}[t] \vspace{-5pt}
    \centering
    \subfloat[]{\label{fig:gyre-distance}\includegraphics[width=0.47\linewidth]{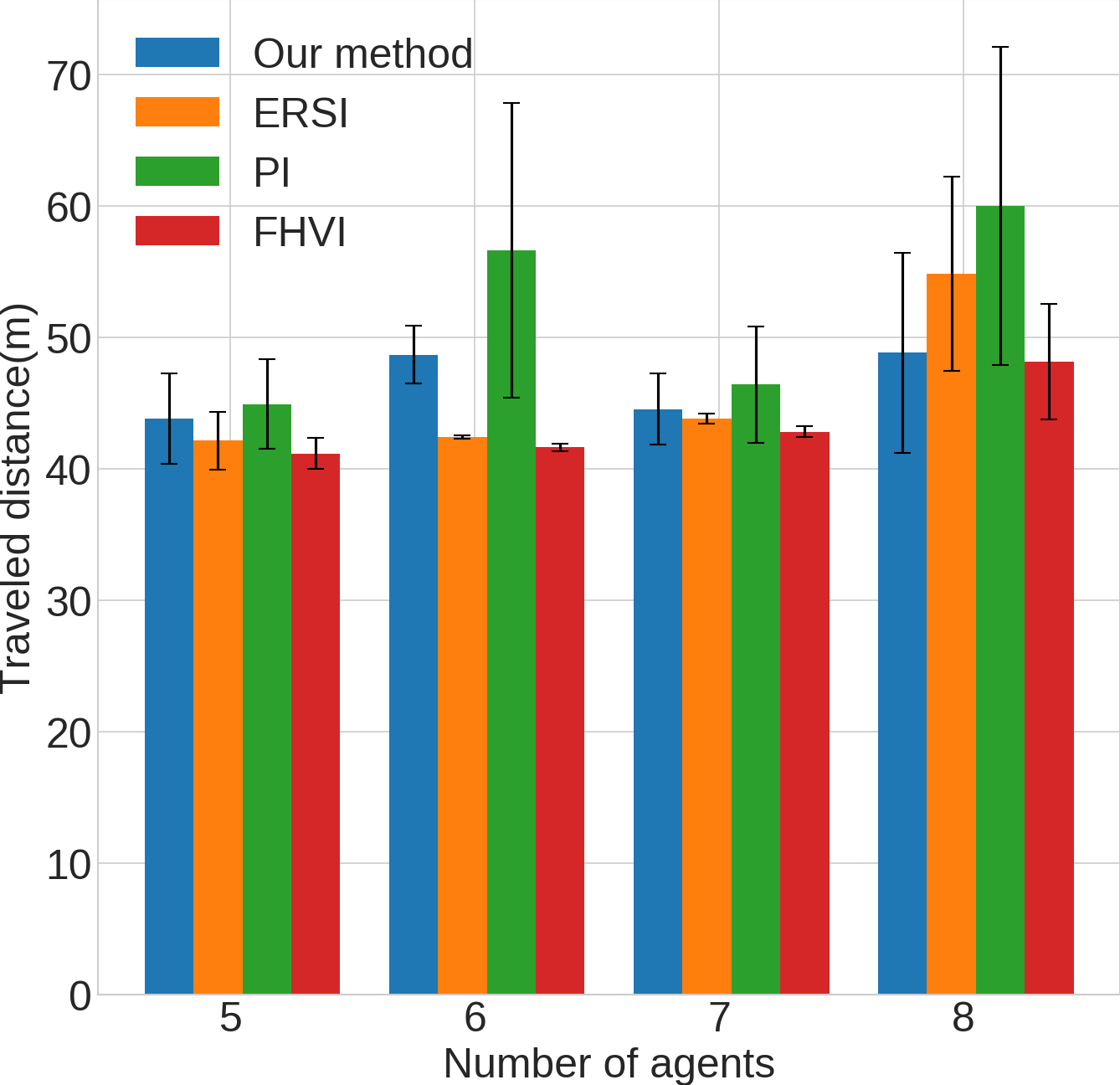}}\quad
    \subfloat[]{\label{fig:gyre-timee}\includegraphics[width=0.47\linewidth]{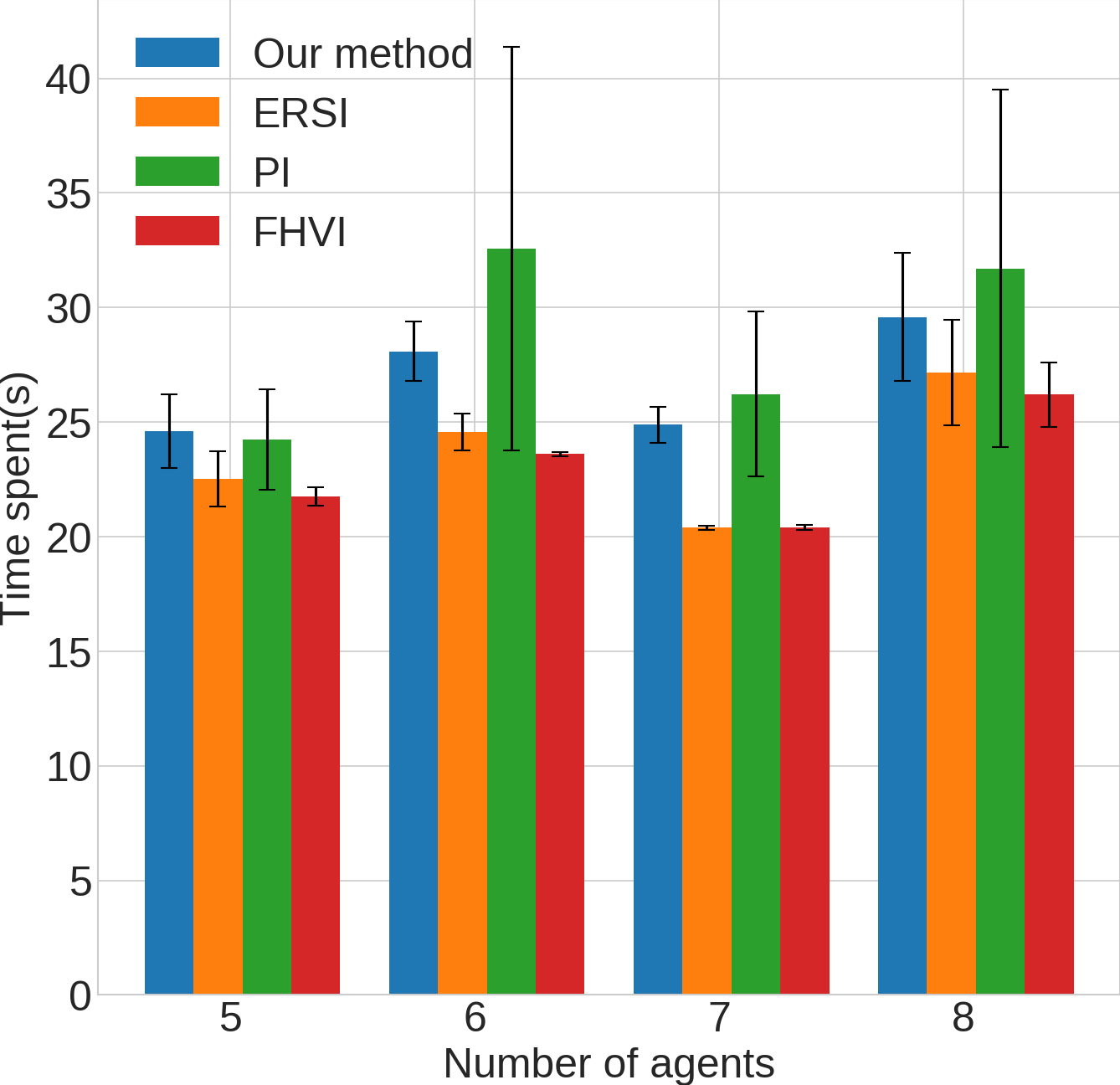}} \vspace{-8pt}\\
    \subfloat[]{\label{fig:vortex-distnace}\includegraphics[width=0.47\linewidth]{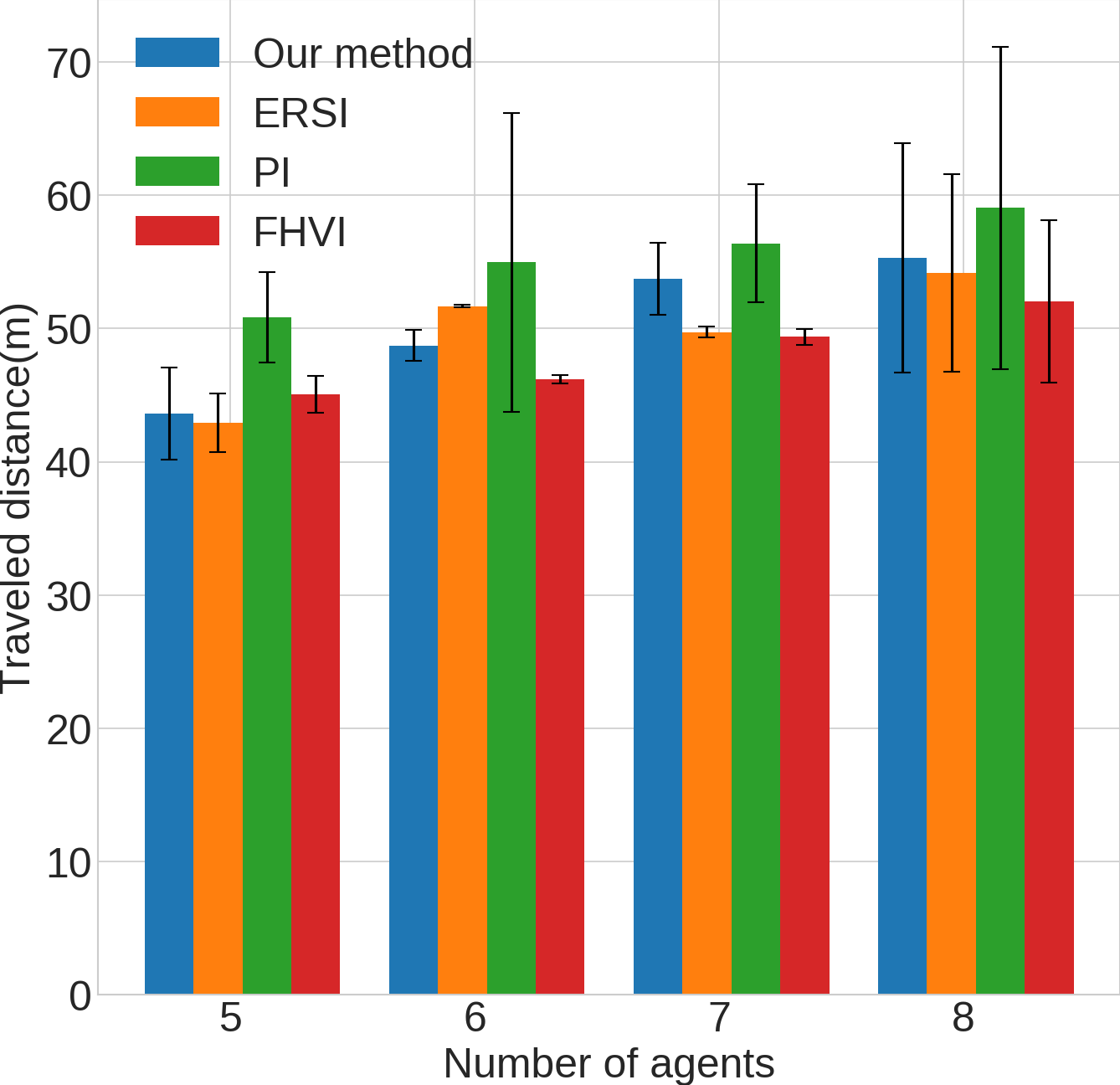}}\quad
    \subfloat[]{\label{fig:vortex-time}\includegraphics[width=0.47\linewidth]{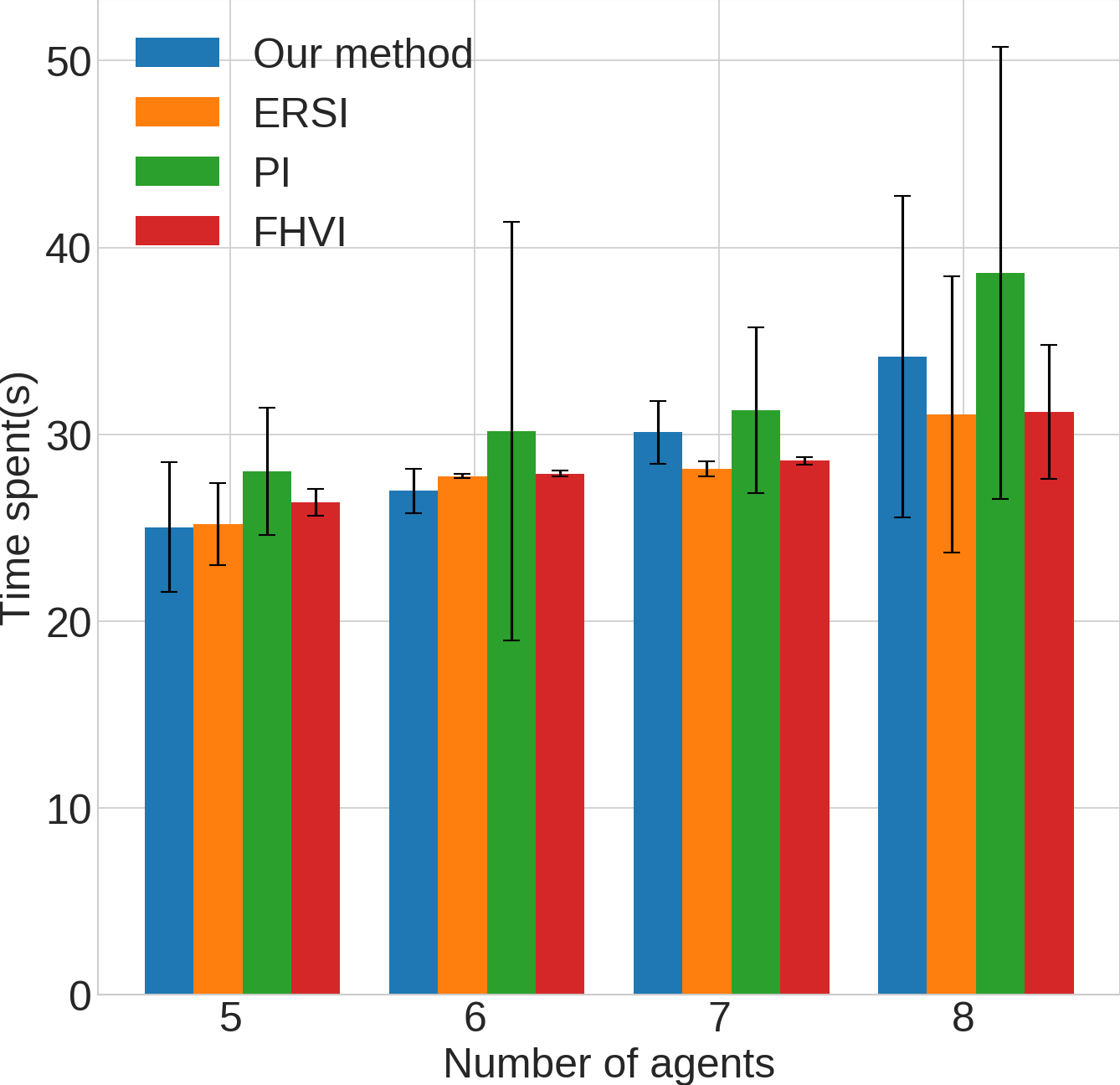}} \vspace{-8pt}
    \caption{\small Results for planning under gyre disturbance (a)(b) and dynamic vortex disturbance (c)(d) of the four algorithms with varying number of dynamic agents in the environment. Each result is averaged over 10 test runs. The first and second column shows the distance traveled and the time taken to the goal, respectively.\vspace{-15pt}
    }
    \label{fig:performance-distance-time}
\end{figure}

\begin{table*}[t]
\caption{Comparison of computational time}
\label{tb:compuation-time}
\centering
\resizebox{0.85\columnwidth}{!}{%
\begin{tabular}{c|cccc|cccc|cccc}
& \multicolumn{4}{c|}{Resolution} & \multicolumn{4}{c|}{State space range}                                                 & \multicolumn{4}{c}{\# of horizon} \\
\hline
& 4m & 2m & 1m & 0.5m & 10m & 20m & 30m & 40m  & 2 & 4 & 6 & 8\\
\hline
FHVI & $0.62s$ & $9.54s$ & $157.25s$ & $2538s$ & 
     $9.43s$ & $157.88s$ & $795.68s$ & $2523s$ & 
     $50.8s$ & $152.8s$ & $254.8s$ & $359.4s$ \\
\hline
ERSI & $0.27s$ & $0.39s$ & $0.96s$ & $3.46s$
   & $0.43s$ & $0.95s$ & $1.85s$ & $3.16s$ 
   & $0.01s$ & $1.0s$ & $80.2$ & $6932s$\\
\hline
PI  & $0.001s$ & $0.01s$ & $0.17s$ & $1.12s$ 
    & $0.12s$ & $0.13s$ & $0.19s$ & $0.19s$ 
    & $0.001s$& $0.17s$ & $0.78s$ & $3.12s$  \\ 
\hline
Ours  & $0.01s$ & $0.02s$ & $0.14s$ & $1.34s$ 
    & $0.13s$ & $0.15s$ & $0.18s$ & $0.20s$ 
    & $0.01s$& $0.14s$ & $0.80s$ & $2.92s$  \\ 
\hline
\end{tabular}
}
\vspace{-5pt}
\end{table*}

We have conducted extensive simulations to validate the proposed method and evaluate the algorithm in terms of the computation time, accuracy, and reliability.

\subsection{General Simulation Setup}
\begin{figure*}[t]     \vspace{-10pt}
\centering
\subfloat[]{\label{fig:}\includegraphics[width=0.23\linewidth]{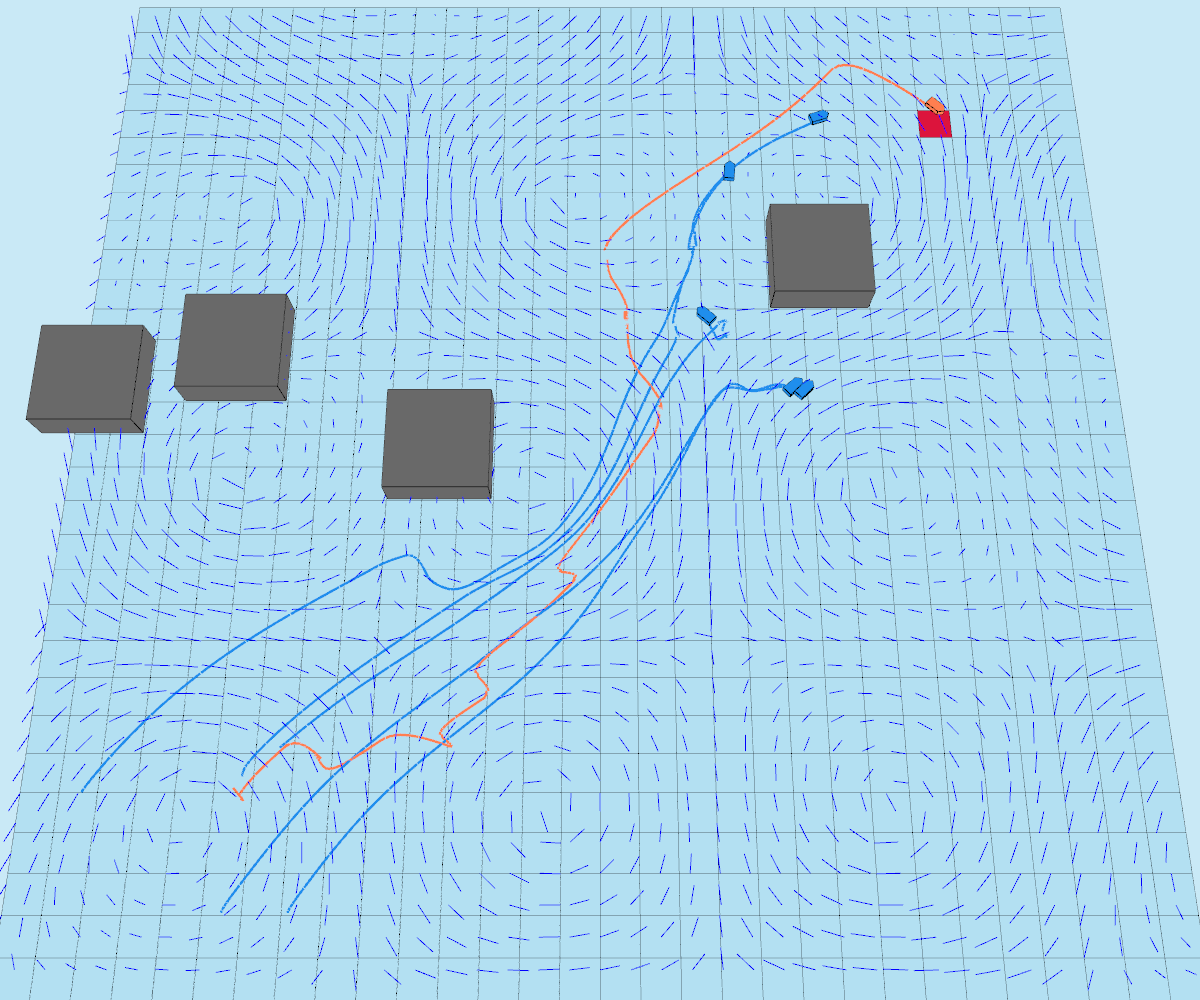}}\quad
\subfloat[]{\label{fig:}\includegraphics[width=0.23\linewidth]{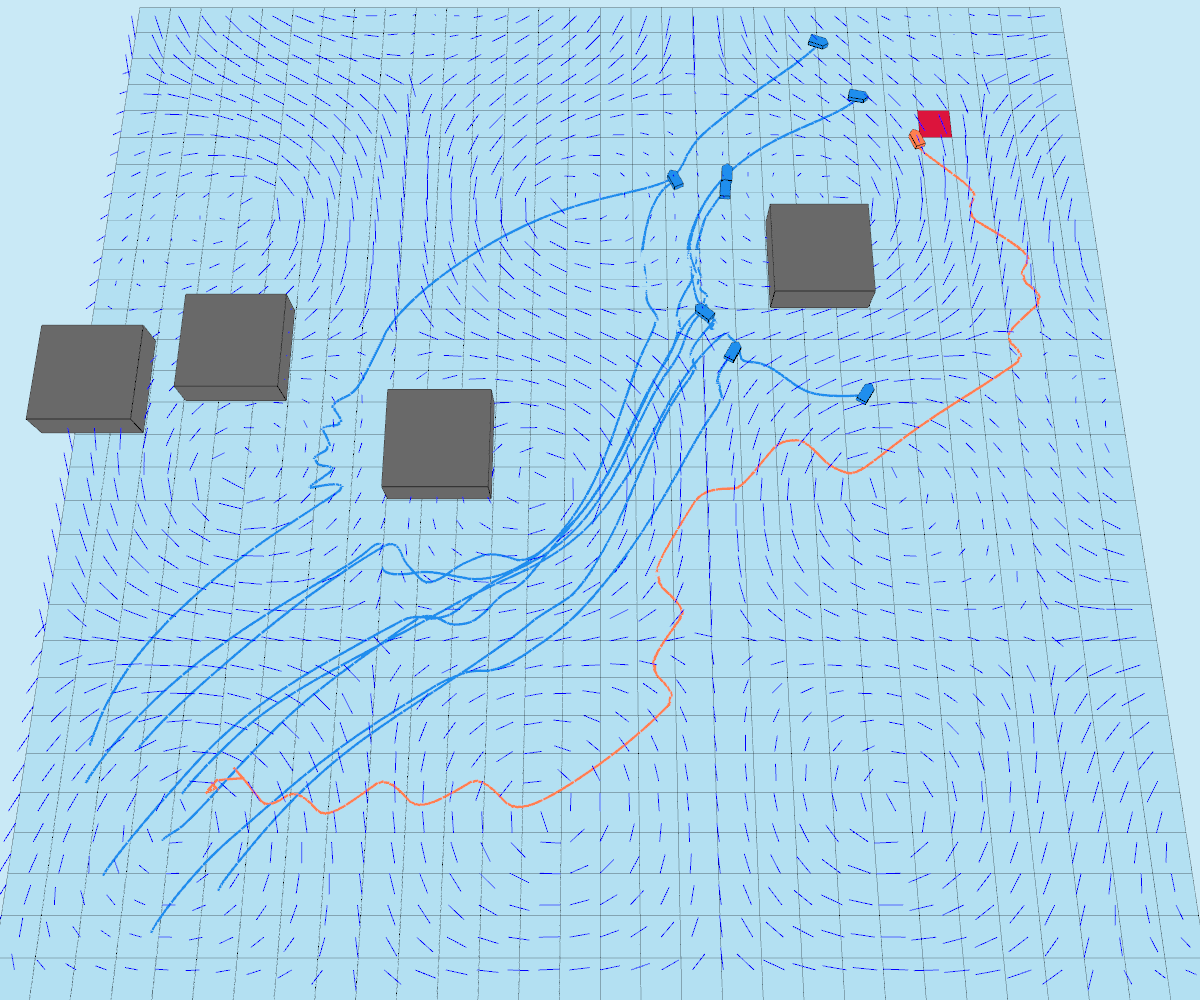}} \quad 
\subfloat[]{\label{fig:}\includegraphics[width=0.23\linewidth]{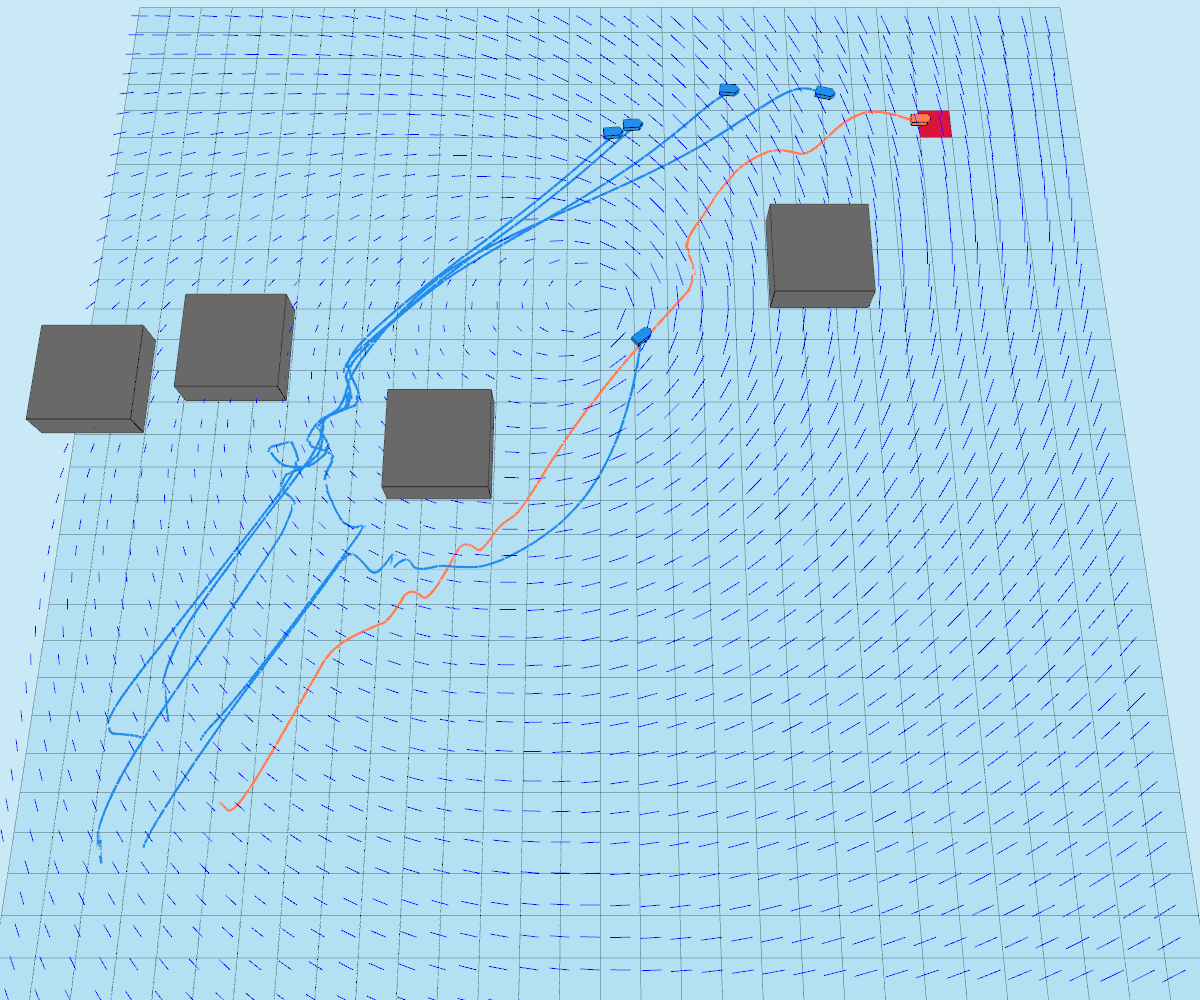}}\quad
\subfloat[]{\label{fig:}\includegraphics[width=0.23\linewidth]{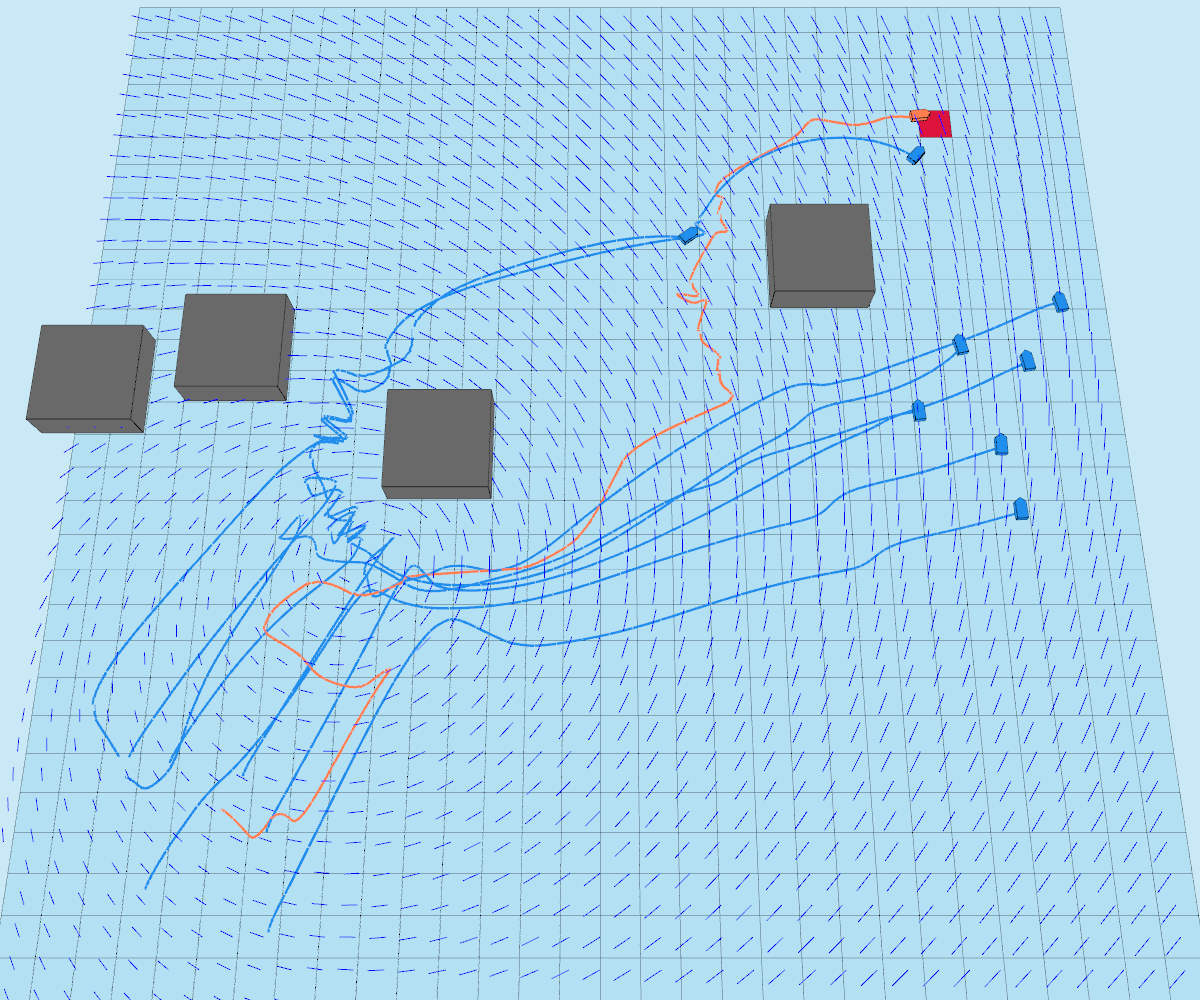}}
\caption{\small Trajectory results of our algorithm in the simulated environment, 
where the trajectories of the controllable and uncontrollable vehicles are shown in blue and orange, respectively. The goal is shown as the red grid. 
The behaviors of other agents are simulated using the social force model which cannot fully cancel (eliminate) external disturbances. 
(a)(b) and (c)(d) show the trajectories under gyre and time-varying vortex disturbances, respectively.
    \vspace{-10pt}
    }
\label{fig:trajectories}
\vspace{-11pt}
\end{figure*}

Our simulation is based on the scenario of marine vehicles that perform tasks on ocean surface.
This scenario is suitable due to the following reasons:
first, both the robot and other vehicles have great action uncertainty due to the time-varying ocean currents; second, traffic and conflict due to multiple vehicles in a limited space can inevitably exist.

The continuous state space is represented by the position of the vehicle $\mathbf{x} = [x, y]^T$. 
The action space consists of velocities $\mathbf{u} = [v_x, v_y]^T$, where both $v_x$ and $v_y$ are within the range of $-2.5m/s$  and $2.5m/s$.
The uncertainty of the environmental disturbance is represented with a $2 \times 2$ identity covariance matrix.
Each dimension of the action space is discretized into $3$ actions with equidistant intervals, so there are $9$ discretized actions in total.

\subsection{Computational Time}
We compare the computation time of our algorithm with other three baseline methods: finite-horizon value iteration (FHVI)~\cite{bertsekas1995dynamic}, exhaustive reachable space iteration (ERSI), and  a modified version of policy iteration (PI).
The PI method improves the policy after evaluating the reachable spaces for all the decision steps, which is used in our previous work~\cite{XuYinLiu2019}. 
ESRI uses all combinations of action sequences to sweep the reachable spaces exhaustively.

Three criteria are designed for evaluating the computational time. 
The results are reported in Table~\ref{tb:compuation-time}.  
The first column analyzes the computation time with respect to different discretization resolutions of the state space.  
We fix the range of the state space $x \in [0m, 20m], y\in [0m, 20m]$ as well as the number of planning horizon $T = 4$.
We set the resolutions of both dimensions to be equal, $\Delta x = \Delta y = h$. 
Thus, the number of states is given by $n = T \times (\frac{20}{h})^2$.
We can observe that, although the computation time grows along with finer resolutions for all algorithms, the reachable space based algorithms are able to solve the problem {\em orders of magnitude}
faster than the FHVI. 

The second column evaluates the computational time with respect to the range of the continuous state space where the resolution and the number of planning horizons are fixed as $h=1m$ and $T=4$.
The numbers in the second row of the second column represent the maximum values of $x$ and $y$ with a minimum value of $0m$. 
The number of discrete states is given by $n = T \frac{l^2}{h}$, where $l$ is the maximum value of the continuous state space.
The results reveal that, if the resolution of each discrete state does not change, the computational time of our algorithm is not affected by increasing the number of discrete states. This also implies that, {\em our method is scalable to the number of states}.

Finally, in the third column, we examine the running time with respect to the number of planning horizons while assuming fixed state ranges and resolutions. 
The results reveal that, 
when the planning horizon is less than $6$ steps, our method and PI are able to finish the computation within 1 second. 
{\em This superior runtime of our method enables the robot to compute online or real-time.} 
In contrast, the ERSI's computational time increases dramatically when the number of horizons increases. 

\subsection{Planning Performance}
\subsubsection{Setup} 
We test our algorithm in a $30m \times 30m$ simulated environment with four static obstacles and $5 - 8$ dynamic vehicles.
The uncontrollable vehicles with SFM behaviors are randomly generated around the robot, 
and the initial condition for each algorithm is the same.
Since the decision processes of other vehicles are not fully observable, we assume the robot holds an inaccurate belief in other vehicles' behaviors, 
i.e., the parameters of the SFM used in predicting other vehicles' behaviors are different from the ones used in the actual experiment simulation.

The spatial resolutions in both $x$ and $y$ dimensions are set to be $1m$. 
All the algorithms plan over four decision steps, i.e., $T=4$, and each action is executed for $\Delta t = 0.5s$. 
We consider the following two types of disturbances:
\begin{enumerate}
    \item Dynamic vortex: the disturbance dynamics is given by
    \begin{equation}
        g(\mathbf{x}, t) = \begin{bmatrix}-\Delta t & 0 \\ 0 & \Delta t \end{bmatrix}\mathbf{x} + \begin{bmatrix}\Delta t & 0 \\ 0 & -\Delta t\end{bmatrix}\mathbf{x}_c(t), 
    \end{equation}
    where the vortex center $\mathbf{x}_c(t) =[r\cos{\omega t} + c_x, r\sin{\omega t}+c_y]^T$ rotates and translates with respect to time.
    $r$ is the rotating radius and $[c_x, c_y]^T$ represents the rotating center. 
    \item Gyre: this is a static and non-linear disturbance which is defined as 
    \begin{equation}
     g(\mathbf{x}, t) = \begin{bmatrix}-\pi A \sin{(\pi \frac{x}{s})}\cos{(\pi \frac{y}{s})\Delta t} \\ \pi A\cos{(\pi \frac{x}{s})}\sin{(\pi \frac{y}{s})\Delta t}\end{bmatrix},
    \end{equation}
    where $A$ is the strength of the disturbance and $s$ determines the size of the gyres.
\end{enumerate}

\begin{figure}[t] \vspace{-10pt}
    \centering
    \subfloat[]{\label{fig:vortex-intervention}\includegraphics[width=0.45\linewidth]{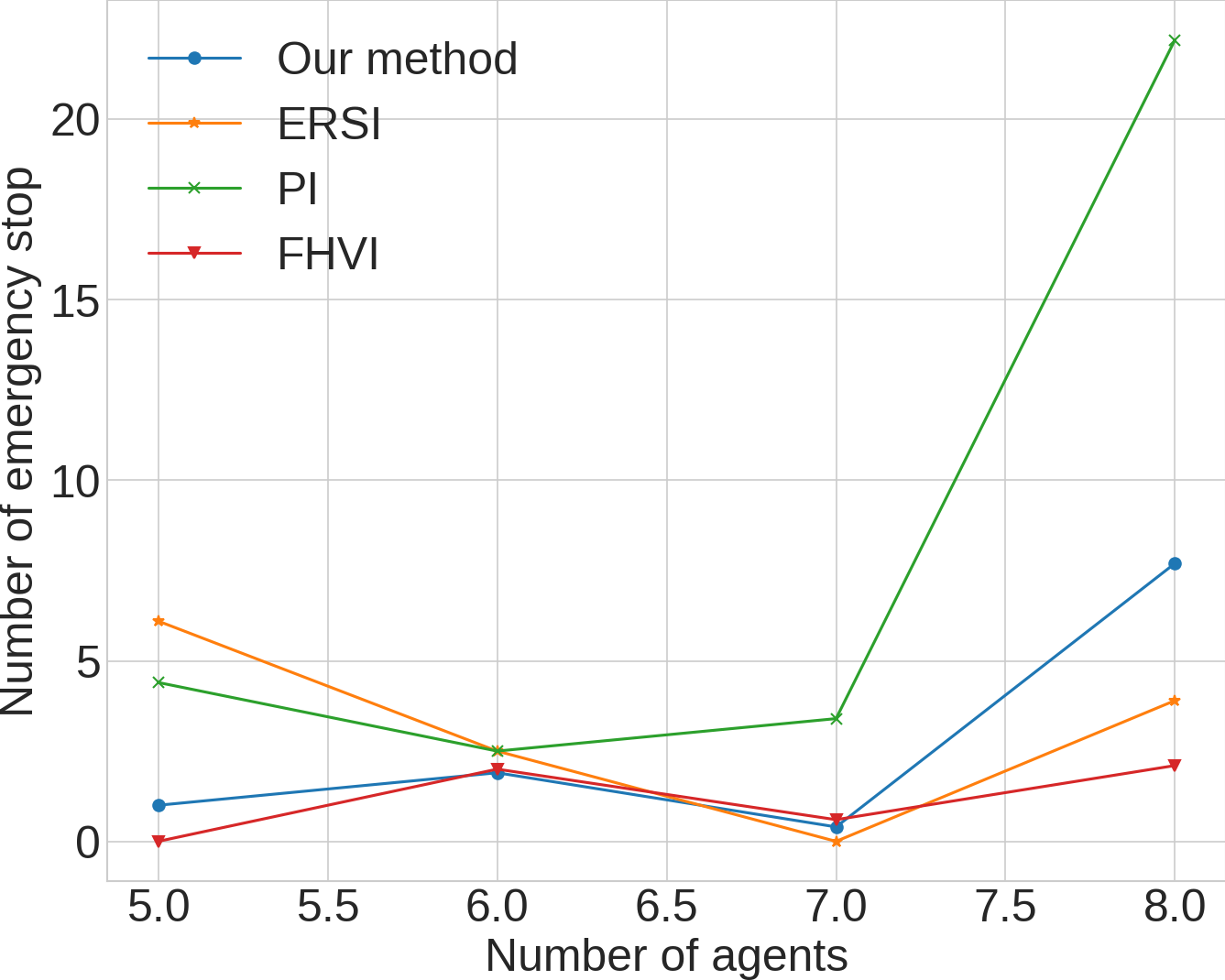}}\quad
    \subfloat[]{\label{fig:vortex-intervention}\includegraphics[width=0.45\linewidth]{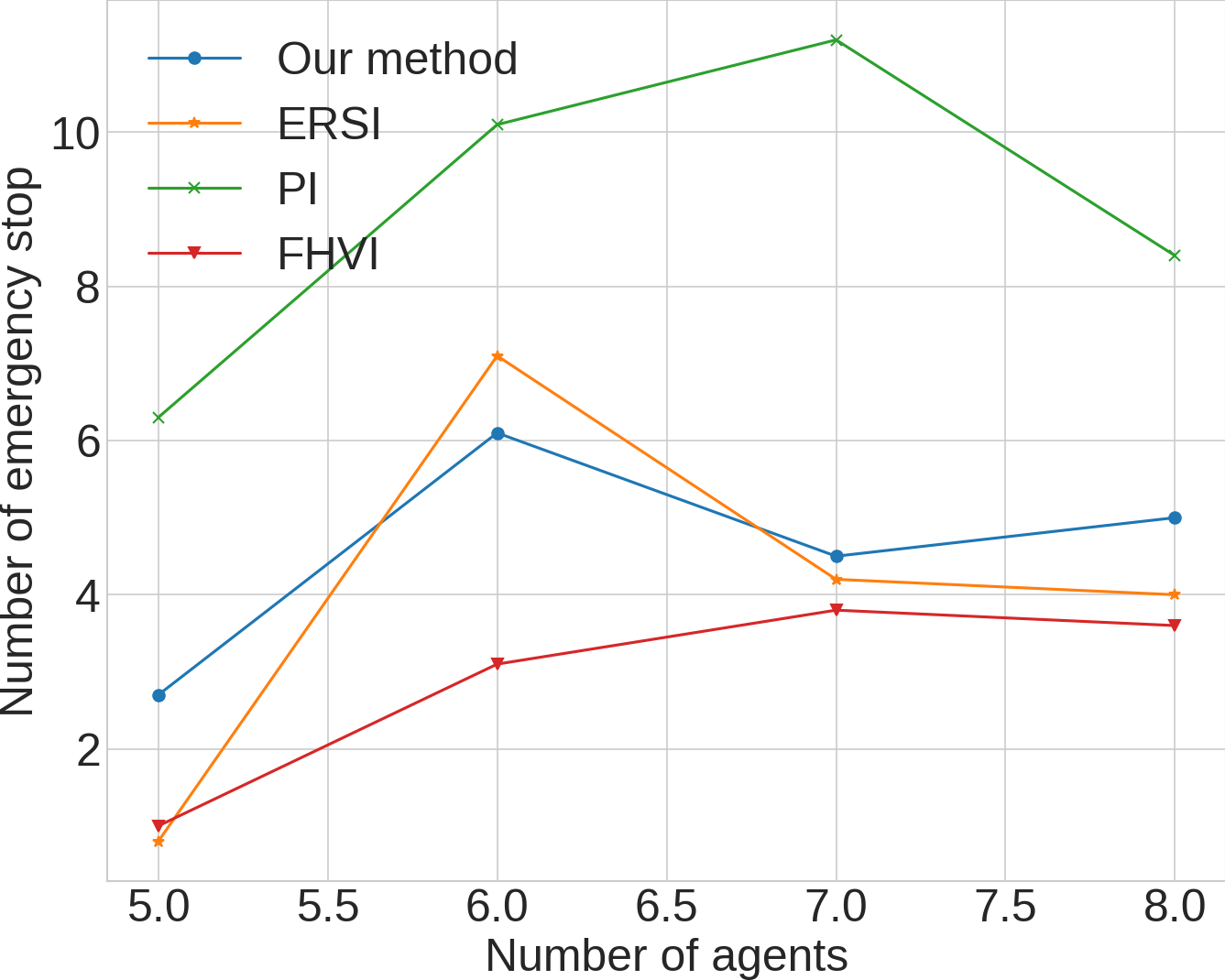}}
    \caption{\small Comparison of the number of emergency stops in (a) gyre disturbance and (b) dynamic vortex disturbance environments.     \vspace{-10pt}}
    \label{fig:performance-emergency-stopping}
    \vspace{-10pt}
\end{figure}

\subsubsection{Accuracy Evaluation}

FHVI provides a performance upper-bound in terms of accuracy (solution optimality) as it exhaustively searches over the whole  state, action, and time spaces.
To achieve online planning, we use a limited computation time budget $0.8s$ for our algorithm and PI.
Note that it is infeasible to use FHVI and ERSI in an online fashion due to their prohibitive computational costs (refer to Table.~\ref{tb:compuation-time}).
To compare our algorithm with these exhaustive search algorithms (ERSI and FHVI), we pause the simulator during their planning phase, and launch all vehicles simultaneously once all planning results are obtained.  

We first compare the distance traveled and the time taken for reaching the goal with different number of agents.
The results are shown in Fig.~\ref{fig:performance-distance-time}.
In general, our algorithm can reach the accuracy that is comparable to the optimal solution from the exhaustive search. It also achieves a remarkably  $10\times$ (compared to ERSI) and $1000\times$ (compared to FHVI) speed-up of the computation time. 
Since the PI method spends most of its computation time calculating the state distributions without improving the policy, the obtained policy is worse than our method within the limited time budget. 

Due to ocean currents, the robot may collide with obstacles or other vehicles. 
For the safety concern, we assume the robot can make an emergency stop before the collision happens.
The more emergency stops, the more unreliable (i.e., unsafe) of a planning method. 
We then record the number of emergency stops needed before arriving at the goal state shown in Fig.~\ref{fig:performance-emergency-stopping}.
The statistics show that the computed policies from our method lead to a number of emergency stops similar to that of the exhaustive search (optimal) algorithms.
Snapshots of trajectories are demonstrated in Fig.~\ref{fig:trajectories}. 


In addition to the above-mentioned evaluations, we compare the proposed method against our previous state discretization based approach~\cite{XuYinLiu2019} using identical settings for fair comparisons.
Specifically, the simulated environment is set with a dimension of $210km \times 234km$ with $\Delta x = \Delta y = 6km$.
The results show that this newly proposed framework uses only $0.4s$ to compute a 4-step look-ahead policy. To obtain a 30-step look-ahead policy, it needs only $4.11s$.
In contrast, the previous method~\cite{XuYinLiu2019} requires 18s to compute a policy with 30 decision steps. 
Its advantageous computational speed allows it to be used in as an online planning algorithm in time-varying dynamic environments.
As shown earlier, this proposed method can also achieve the planning accuracy comparable to the optimal solution.


\vspace{-5pt}
\section{Conclusion}
This paper presents an online TVMDP-based algorithm to solve the robot navigation problem in a time-varying uncertain environment in the presence of other mobile vehicles. 
Viewing this problem as a TVMDP has allowed us to naturally introduce the dynamic-obstacle-aware reachable space based on the confidence region of robot's spatial state distribution to reduce the computational time. 
Moreover, we embed the ideas of unscented transform 
to remarkably improve the estimation accuracy of the spatial state distribution with nonlinear vehicle dynamics. 
Extensive simulation experiments have revealed significant advantages of this proposed new framework in terms of computational time, decision accuracy, and planning reliability. 

\vspace{-3pt}
{\small
\bibliographystyle{IEEEtran}
\bibliography{ref}
}
\end{document}